\documentclass{article}
\usepackage{spconf,amsmath,graphicx}

\usepackage{booktabs}
\usepackage{algorithmic}
\usepackage[linesnumbered,ruled,vlined]{algorithm2e}
\SetKwInput{KwInput}{Input}                
\SetKwInput{KwOutput}{Output}              

\usepackage{amsmath}
\usepackage{graphicx}
\usepackage{subcaption}
\usepackage{setspace}
\usepackage{color}

\usepackage{paralist}

\title{Synergistic Network Learning and Label Correction for Noise-robust Image Classification}
\name{Chen Gong\textsuperscript{$\dagger$}\textsuperscript{1}, Kong Bin\textsuperscript{$\dagger$}\textsuperscript{2}, Eric J. Seibel\textsuperscript{1}, Xin Wang{$^*$}\textsuperscript{2}, Youbing Yin{$^*$}\textsuperscript{2}, Qi Song\textsuperscript{2} \thanks{\textsuperscript{$\dagger$} Authors contributed equally to this work. $^*$ Corresponding authors. }}
\vspace{-0.6cm}
\address{\textsuperscript{1}Mechanical Engineering, University of Washington, Seattle, WA 98195, USA\\
\textsuperscript{2}Keya Medical, Seattle, WA 98104, USA
}
\vspace{-1cm}
\begin{document}
%
\maketitle
\begin{abstract}
\vspace{-0.1cm}
Large training datasets almost always contain examples with inaccurate or incorrect labels. Deep Neural Networks (DNNs) tend to overfit training label noise, resulting in poorer model performance in practice. To address this problem, we propose a robust label correction framework combining the ideas of small loss selection and noise correction, which learns network parameters and reassigns ground truth labels iteratively.
Taking the expertise of DNNs to learn meaningful patterns before fitting noise, our framework first trains two networks over the current dataset with small loss selection. Based on the classification loss and agreement loss of two networks, we can measure the confidence of training data. More and more confident samples are selected for label correction during the learning process.
We demonstrate our method on both synthetic and real-world datasets with different noise types and rates, including CIFAR-10, CIFAR-100 and Clothing1M,
where our method outperforms the baseline approaches.
\end{abstract}
\vspace{-0.4cm}
\section{Introduction}
\vspace{-0.4cm}
Deep learning has shown very impressive performance on various vision problems. However, a practical challenge for deep learning state-of-the-art models is that they rely on large amounts of clean, annotated data \cite{krizhevsky2012imagenet,he2016deep, Li2017WebVisionDV, mahajan2018exploring}. 
Collecting such data sets is expensive or time-consuming. Large training datasets almost always contain examples with inaccurate or incorrect labels, resulting in overfitting to noisy samples and poorer model performance \cite{frenay2013classification,hu2020learning,hu2021tkml,hu2021sum}. There are many classic methods to prevent noise overfitting such as dropout \cite{srivastava2014dropout} and early stop \cite{zhang2005boosting}, which are heuristic for noisy label learning.

A large number of algorithms have been developed for learning with noisy labels. Small loss selection recently achieved great success on noise-robust deep learning following the widely used criterion: DNNs tend to learn simple patterns first, then gradually memorize all samples \cite{krueger2017closer, 9070553, zhang2021understanding}. These methods treat samples with small training loss as clean ones. During the training process, clean or confident instances are selected to update the model parameters. 
For instance, MentorNet \cite{jiang2018mentornet} trains a Teacher-Net to provide curriculum, in the form of weights on the training samples,  to select clean samples to guide the training of the Student-Net. 
Co-teaching \cite{han2018co} uses two networks to determine clean samples in their mini-batches separately and exchange the update information with the other network. Inspired from Co-teaching, JoCoR \cite{jocor} also uses two networks, while they combine the "agreement strategy"  from semi-supervised learning into noise label tasks, which uses a joint loss to make their predictions agree. 
The instances with small joint loss are selected for the back-propagation. Experiments in JoCoR showed that it is more effective than co-teaching.  
However, not all clean examples can be selected by the network with the small loss strategy. When the noise rate is high, the selection would further decrease the number of effective training samples in each batch. 
Currently, there are relabelling methods learning network parameters and inferring the ground-truth labels simultaneously without any clean dataset, such as joint optimization \cite{tanaka2018joint} and PENCIL \cite{yi2019probabilistic}. They use all data for learning so noisy examples can be an interference in this process.

\begin{figure}[t]
\centering
\includegraphics[scale=0.35]{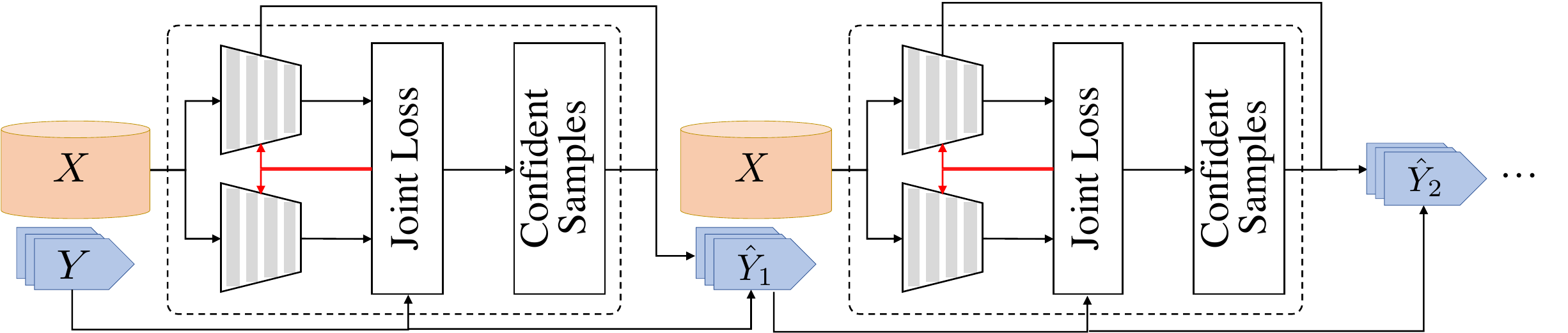}
\caption{The iterative network learning and label correction of proposed method. Two models are trained with a joint loss and confident samples are selected for relabelling.}
\label{fig:overview}
\vspace{-0.6cm}
\end{figure}

In this paper, motivated by increasing the effective training samples, we proposed a framework that combines the noise correction with small loss selection methods to update the network parameters and noisy labels iteratively (Fig. \ref{fig:overview}). Each time we train relatively robust models with small loss examples, and correct the noisy labels which the current networks are confident on. In the training process, we are inspired by JoCoR to use "agreement strategy" to train two networks with a joint loss. Our contributions are as follows.

(a) We proposed a robust learning framework for both network parameters and ground truth labels to handle noise label tasks. Our method is independent of the backbone network structure and does not need an auxiliary clean dataset.
Training the network with small loss selection and updating the label of confident examples make the iterative learning process robust. To the best of our knowledge, it is the first method in this line. 
(b) We prove the effectiveness of training a stable model over current dataset before each label correction step. It performs better than conducting weights and label update at each iteration.
(c) We conduct extensive experiments on both synthetic and real-world noisy datasets and our method achieves state-of-the-art accuracy.

\vspace{-0.4cm}
\section{Proposed Approach}


\vspace{-0.4cm}
\subsection{Notation}
For multi-class classification with $C$ classes, we have a dataset $D$ with $N$ samples. $D = \left\{x_i, y_i\right\}_{i=1}^N$, where $x_i$ is the $i$th example with its label as $y_i \in [1, . . . ,C]$.  The set of examples and labels are denoted as $X$ and $Y$.
The neural network is denoted by $f(x,\theta)$, with the output of the final layer (C-class softmax layer): ${\bf p} = [p_1,...,p_C]$. ${\bf p}$ is the predicted probability for each example $x$.

During network training in a standard classification task, a loss function $\mathcal{L}$ is used to measure the distance between ${\bf p}$ and the label $y$.
The network parameters $\theta$ are learned by optimizing $\mathcal{L}$ by gradient descent methods. 
In our study, the given label $Y$ is noisy, and our task is to jointly optimize $\theta$ and labels $Y$. The learned label set is denoted as $\hat{Y}$.

\vspace{-0.4cm}
\subsection{Joint Training with Small Loss Selection}
According to the agreement maximization principles \cite{kumar2010co,zhang2018deep,jocor}, different models would agree on labels of most examples and are unlikely to agree on incorrect labels. In the process of updating $\theta$, two different classifiers are encouraged to make predictions closer to each other with a regularization term to reduce divergence. 
Specifically, joint training uses a joint loss to train two networks (same structure with different initialization) simultaneously. 
The loss function is composed of supervised learning loss and agreement loss as shown in Eq. \ref{joint},
\begin{equation} \label{joint}
\mathcal{L}_{joint} = (1-\lambda)*\mathcal{L}_{sup}+\lambda*\mathcal{L}_{agr},
\end{equation}
where $\lambda$ is a hyperparameter for linear combination and it decreases with the label noise rate.

The supervised learning loss $\mathcal{L}_{sup}$ is the sum of the classification loss of two networks, which is the Cross-Entropy between predictions and labels. 
The agreement loss is the term to reduce divergence between two classifiers which is the Jensen-Shannon divergence in Eq. \ref{eq:lagr}, where $\mathcal{D}_{KL}(p_1||p_2)$ is the Kullback–Leibler divergence.  
\begin{equation} \label{eq:lagr}
\mathcal{L}_{agr} = \mathcal{D}_{KL}(p_1||p_2) + \mathcal{D}_{KL}(p_2||p_1),
\end{equation}
Small loss examples are selected to do the back-propagation in each iteration and a selection rate $R(t)$ is used to control the portion. $R(t)$ depends on the iteration $t$ and its schedule is introduced in Section 2.4. 


\vspace{-0.4cm}
\subsection{Learning with label correction}
\vspace{-0.2cm}
From the view of agreement maximization principles that different models are unlikely to agree on incorrect labels, the joint loss would be high with both high supervised learning loss and agreement loss over noisy labelled examples. 
Given this assumption, we obtain clean labels by updating labels in the direction to decrease Eq. \ref{min}.
\begin{equation} \label{min}
\min_{\theta_1,\theta_2,Y} \mathcal{L}_{joint}(\theta_1,\theta_2,Y|X).
\end{equation}
In the proposed learning framework, network parameters $\theta_1, \theta_2$ and class labels $Y$ are alternatively updated:
1) Updating $\theta$ with fixed $Y$: With the joint loss in Eq. \ref{joint}, we update $\theta_1$ and $\theta_2$ with small loss selection.
2) Updating $Y$ with fixed $\theta_1, \theta_2$: After relatively robust models are trained on the current dataset, part of the noisy labels are corrected.

When updating the labels $Y$, we want to select the labels which are noisy and the networks are confident to update. 
The examples with small agreement loss are considered as confident examples so we select the subset $D_{correction}$ with the least agreement loss based on the label correction rate:
\begin{equation} \label{D1}
D_{confident} = argmin_{D':|D'|\geq C(k)|D|}\mathcal{L}_{agr}(D'),
\end{equation}
where $C(k)$ is the label correction rate which changes with the $k^{th}$ label correction. 
Among the confident examples, we take the noisy ones whose predicted class distributions have large difference with the current labels. 
Similarly, this subset $D_{noisy}$ can be selected with the large supervised learning loss according to the label correction rate $C(k)$:
\begin{equation} \label{D2}
D_{noisy} = argmax_{D':|D'|\leq C(k)|D|}\mathcal{L}_{sup}(D').
\end{equation}
The subset for label correction $D_{correction}$ will be the intersection of $D_{noisy}$ and $D_{confident}$:
\begin{equation} \label{eq:D}
D_{correction} = D_{confident}\cap D_{noisy}.
\end{equation}
Within the selected examples, if two networks have the same predictions on one example, the current label is updated with the predicted class, otherwise it remains unchanged.

Instead of updating labels and parameters simultaneously in each iteration, we train two robust models on the current dataset before every label update step to make the correction more reliable. 
Since labels are not continuously updated, there can be a large amount of changes on the dataset after correction. We set a threshold $C_{restart}$ for label correction rate $C(k)$ as a trigger for restarting training. When $C(k)>C_{restart}$, a large portion of labels will be changed and we retrain two networks from scratch with the new labelled dataset.

We can use any deep neural network as the backbone network, and then equip it with the joint training and label correction to handle learning problems with noisy labels. After the networks have been fully trained, the label correction parts are not needed. The backbone networks alone can perform prediction for test examples and indicates ambiguous ones when two networks disagree.

\vspace{-0.4cm}
\subsection{Training implementation}
The label correction rate $C(k) = \tau/2*1/k$ since as the number of correction time $k$ increases, fewer examples are noise-labelled. The raw noise ratio $\tau$ decreases after label update according to the size of last $D_{correction}$:
\begin{equation} \label{eq:tau}
\tau_k = \tau_{k-1} - \frac{|D_{correction}|}{|D|}.
\end{equation}
The schedule of $R(t) = 1 - min\left\{\frac{t}{T_k}\tau, \tau \right\},$ where $t$ is the iteration and $T_k = 10$ for CIFAR-10 and CIFAR-100, $T_k = 5$ for Clothing1M.

The training of our method is implemented through two stages: (1) Iterative optimization of network parameters $\theta_1, \theta_2$ and labels $\hat{Y}$. (2) Fine-tuning $\theta_1$ and $\theta_2$ with fixed labels $\hat{Y}$.

\vspace{-0.4cm}
\section{Experiments}

\vspace{-0.4cm}
\subsection{Datasets}

Our method is demonstrated on synthetic datasets CIFAR-10, CIFAR-100, and noisy dataset Clothing1M \cite{xiao2015learning}, using the PyTorch framework. We corrupted synthetic datasets with symmetric and asymmetric noise manually. The asymmetric noise is to simulate that labellers may make mistakes only within very similar classes. 

A label transition matrix $Q$ is used to flip clean label $y$ to noise label $\hat{y}$ \cite{reed2014training, patrini2017making}. In the symmetric noise setup, label noise is uniformly distributed among all categories, and the label noise ratio is $\tau \in [0, 1]$. For each example, the noise-contaminated label has $1-\tau$ probability to remain correct, but has $\tau$ probability to be flipped uniformly with other $c-1$ labels. All datasets have the same symmetric noise setting.
As for asymmetric noise, 
The noisy labels of CIFAR-10 were generated by mapping $truck \rightarrow automobile$, $deer \rightarrow horse$, $bird \rightarrow airplane$ and $cat \leftrightarrow dog$ \cite{yi2019probabilistic} with probability $\tau$. In CIFAR-100, 100 classes are grouped into 20 5-size superclasses and each class is flipped into the next circularly with noise ratio $\tau$ within each superclass \cite{jocor}.

Clothing1M is a large-scale dataset from 14 clothing classes with noisy labels. The estimated noise level is 40\% \cite{xiao2015learning}.
we use the 1M images with noisy labels for training, the 14k and 10k clean data for validation and test, respectively. We resize the image to $256*256$ and crop the middle $224*224$ as input, then perform normalization.
This dataset is seriously imbalanced among different classes and the label mistakes mostly happen between similar classes. PENCIL \cite{yi2019probabilistic} randomly selected a small balanced subset to relieve the difficulty caused by imbalance. 
\vspace{-0.4cm}
\subsection{Implementation details}
We use a 7-layer CNN network architecture for CIFAR-10 and CIFAR-100 and 18-layer ResNet for Clothing1M, which are same in JoCoR for fair comparison. 

For experiments on CIFAR-10 and CIFAR-100, Adam optimizer with momentum=0.9 is used and the batch size is 128. An initial learning rate of 0.001 is used in the first 250 epochs for iterative parameter and label learning. For fine-tuning the network with fixed labels, we run 50 epochs and linearly decay the learning rate from 0.001 to zero. 
The $\lambda$ in Eq. \ref{joint} balances the supervised learning loss and agreement loss. We linearly decrease $\lambda$ from 0.9 to 0.7 after each label update and keep 0.7 at the fine-tuning stage.
The label update interval is 50 epochs. The label correction threshold $C_{retrain} = 5\%$.

In Clothing1M, the same Adam optimizer is used and the batch size is 64. We run 15 epochs in iterative optimization and 10 epochs for fine-tuning. The learning rate for the first 5 epochs is $8\times10^{-4}$, second 10 epochs is $5\times10^{-4}$, then for fine-tuning is $5\times10^{-5}$.
$\lambda$ is set to $0.85$. $T_{update} = 5$ and the retraining threshold $C_{retrain} = 5\%$.

\vspace{-0.4cm}
\subsection{Comparison with baselines}
\vspace{-0.2cm}
We compare our method with following baseline algorithms on each dataset: F-correction \cite{patrini2017making}, Co-teaching \cite{han2018co}, Co-teaching+ \cite{coteaching+}, JoCoR \cite{jocor}, PENCIL \cite{yi2019probabilistic}, DivideMix (Clothing1M dataset) \cite{li2020dividemix} and $L_{DMI}$ \cite{xu2019l_dmi}. 
The performance of standard deep networks training on noisy datasets is also used as a simple baseline. The performance of PENCIL, DivideMix and $L_{DMI}$ is based on our implementation, and the others are quoted from \cite{jocor}.

{\bf CIFAR-10:}
The test accuracy and standard deviation compared with baselines over CIFAR-10 are shown in Table. \ref{tb:cifar10}.
Our method performs the best in all four cases especially with symmetric-50\% noise (+6.21).
We can see as the noise ratio increased, the standard deviation increases which indicates the stability of models is affected by noise labels. Our method has a smaller standard deviation especially in the high noise rate. Our standard deviation is 0.89 in symmetry-80\%, while that of the other top-3 methods co-teaching, JoCoR and PENCIL are 2.22, 3.06 and 1.86, respectively. 

\begin{table*}[!htbp]
\centering
	\vspace{-0.4cm}
\caption{Average test accuracy (\%) and standard deviation on CIFAR-10 of 5 trials.}
	\vspace{-0.4cm}
\resizebox{\linewidth}{!}{
\begin{tabular}{c|ccccccc|c}
\toprule 
\hline
Flipping Rate & Standard & F-correction & Co-teaching & Co-teaching+ & JoCoR & PENCIL & $L_{DMI}$ & Ours\\
\hline
Symmetry-20\% & 69.18 $\pm$ 0.52 & 68.74 $\pm$ 0.20 & 78.23 $\pm$ 0.27 & 78.71 $\pm$ 0.34 & 85.73 $\pm$ 0.19 & 81.35 $\pm$ 0.32  & 80.52 & \textbf{86.84 $\pm$ 0.18}\\
\hline
Symmetry-50\% & 42.71 $\pm$ 0.42 & 42.19 $\pm$ 0.60 & 71.30 $\pm$ 0.13 & 57.05 $\pm$ 0.54 & 79.41 $\pm$ 0.25 & 69.29 $\pm$0.78 & 73.21& \textbf{85.62 $\pm$ 0.48}\\
\hline
Symmetry-80\% & 16.24 $\pm$ 0.39 & 15.88 $\pm$ 0.42 & 26.58 $\pm$ 2.22 & 24.19 $\pm$ 2.74 & 27.78 $\pm$ 3.06 & 25.30$\pm$ 1.86 & 21.48& \textbf{29.01 $\pm$ 0.89} \\
\hline
Asymmetry-40\% & 69.43 $\pm$ 0.33  & 70.60 $\pm$ 0.40  & 73.78 $\pm$ 0.22  & 68.84 $\pm$ 0.20  & 76.36 $\pm$ 0.49
  & 68.53$\pm$0.48  & 65.12 &\textbf{76.83 $\pm$ 0.29} \\
  \hline
\bottomrule
\end{tabular}}
\label{tb:cifar10}
\vspace{-0.4cm}
\end{table*}

{\bf CIFAR-100:}
we compare our results with other baselines on CIFAR-100 in Table \ref{tb:cifar100}. CIFAR-100 has a similar dataset size to CIFAR-10 while with 10 times of classes, thus the classification is more challenging. Our method is still the overall accuracy winner. In symmetry-20\% and symmetry-50\%, Our method and JoCoR works significantly better than other methods and ours has an advantage of $+1.6\sim1.9$ over JoCoR. In the hardest case of symmetry-80\%, JoCoR ties together with Co-teaching and ours has $+3.36$ test accuracy. The standard classifier, F-correction and PENCIL fail in this case with the accuracy below 5\%. In terms of asymmetry-40\% noise, Co-teaching+ performs better than other baselines with 33.62\% accuracy whereas that of ours is 34.41\%.

\begin{table*}[!htbp]
\centering
\caption{Average test accuracy (\%) and standard deviation on CIFAR-100 of 5 trials.}
	\vspace{-0.4cm}
\resizebox{\linewidth}{!}{
\begin{tabular}{c|ccccccc|c}
\toprule 
\hline
Flipping Rate & Standard & F-correction & Co-teaching & Co-teaching+ & JoCoR & PENCIL & $L_{DMI}$& Ours\\
\hline
Symmetry-20\% & 35.14 $\pm$ 0.44 & 37.95 $\pm$ 0.10 & 43.73 $\pm$ 0.16 & 49.27 $\pm$ 0.03 & 53.01 $\pm$ 0.04
 & 43.61 $\pm$0.23 & 48.95 & \textbf{54.90 $\pm$ 0.07}\\
\hline
Symmetry-50\% & 16.97 $\pm$ 0.40 & 24.98 $\pm$ 1.82 & 34.96 $\pm$ 0.50 & 40.04 $\pm$ 0.70 & 43.49 $\pm$ 0.46 & 26.41 $\pm$0.51 & 39.21 & \textbf{45.12 $\pm$ 0.42}\\
\hline
Symmetry-80\% & 4.41 $\pm$ 0.14 & 2.10 $\pm$ 2.23 &15.15 $\pm$ 0.46 & 13.44 $\pm$ 0.37 & 15.49 $\pm$ 0.98
  & 3.65$\pm$0.77 & 10.65 & \textbf{18.85 $\pm$ 0.59} \\
\hline
Asymmetry-40\% & 27.29 $\pm$ 0.25  & 25.94 $\pm$ 0.44   & 28.35 $\pm$ 0.25    & 33.62 $\pm$ 0.39   & 32.70 $\pm$ 0.35  & 27.32 $\pm$0.42  & 31.34 & \textbf{34.41 $\pm$ 0.72 } \\
\hline
\bottomrule
\end{tabular}}
\label{tb:cifar100}
\vspace{-0.4cm}
\end{table*}



{\bf Clothing1M:}
We demonstrate our method on real-world noisy labels with Clothing1M dataset, which includes a lot of unknown structure (asymmetric) noise. The results are shown in Table. \ref{tb:c1m}, where “Best” denotes the test accuracy of the epoch where the validation accuracy is optimal and “last” denotes the test accuracy of the last epoch. 
We use the complete noisy training set of Clothing1M instead of the small pseudo-balanced subset.
Our method achieved SOTA performance in "Best". In the last epoch, our performance is slightly decreased because of overfitting and $0.05\%$ lower than $L_{DMI}$. The accuracy drop can be improved by stopping training early according to the validation set in practice. 

\begin{table}[t]
\footnotesize
\centering
\caption{Average accuracy (\%) on Clothing1M test set.}
	\vspace{-0.4cm}
\setlength{\tabcolsep}{0.04\textwidth}{
\begin{tabular}{c|c|c}
\toprule 
\hline 
Methods & Best & Last\\
\hline
Standard&67.22 & 64.68\\
\hline
F-correction \cite{patrini2017making}  &68.93&65.36 \\
\hline
Co-teaching \cite{han2018co} &69.21&68.51\\
\hline
Co-teaching+ \cite{coteaching+} &59.32&58.79\\
\hline
JoCoR \cite{jocor}&70.30&69.79\\
\hline
PENCIL \cite{yi2019probabilistic} & 69.48 & 68.48\\
\hline
$L_{DMI} \cite{xu2019l_dmi}$ & 70.79 & \textbf{70.02}\\
\hline
DivideMix \cite{li2020dividemix} & 68.24 & 67.19\\
\hline
Ours & \textbf{71.63} & 69.97\\
\hline
\bottomrule
\end{tabular}}
\label{tb:c1m}
	\vspace{-0.4cm}
\end{table}

\vspace{-0.4cm}
\subsection{Ablation Study}
We conduct ablation study for analyzing the effect of label update interval and retraining. The experiments are set up on CIFAR-10 with Symmetry-50\% noise.

{\bf Label Update Interval:}
Instead of referring clean labels in each iteration \cite{tanaka2018joint,yi2019probabilistic}, our label correction is intermittent. With label update intervals, two classifiers can be trained on the current dataset with small loss selection strategy and try to maximize the use of current data. The agreement maximization principle will be more reliable with these relatively robust models. 
In this experiment, we set the label update interval to 50 epochs and compared it with the continuous update in each iteration. 
Similar to Joint optimization \cite{tanaka2018joint}, a pre-trained backbone is needed for continuous label update. We train two networks for 50 epochs before the label update both in continuous and intermittent methods.

\begin{figure}[t]
\centering
\includegraphics[scale=0.45]{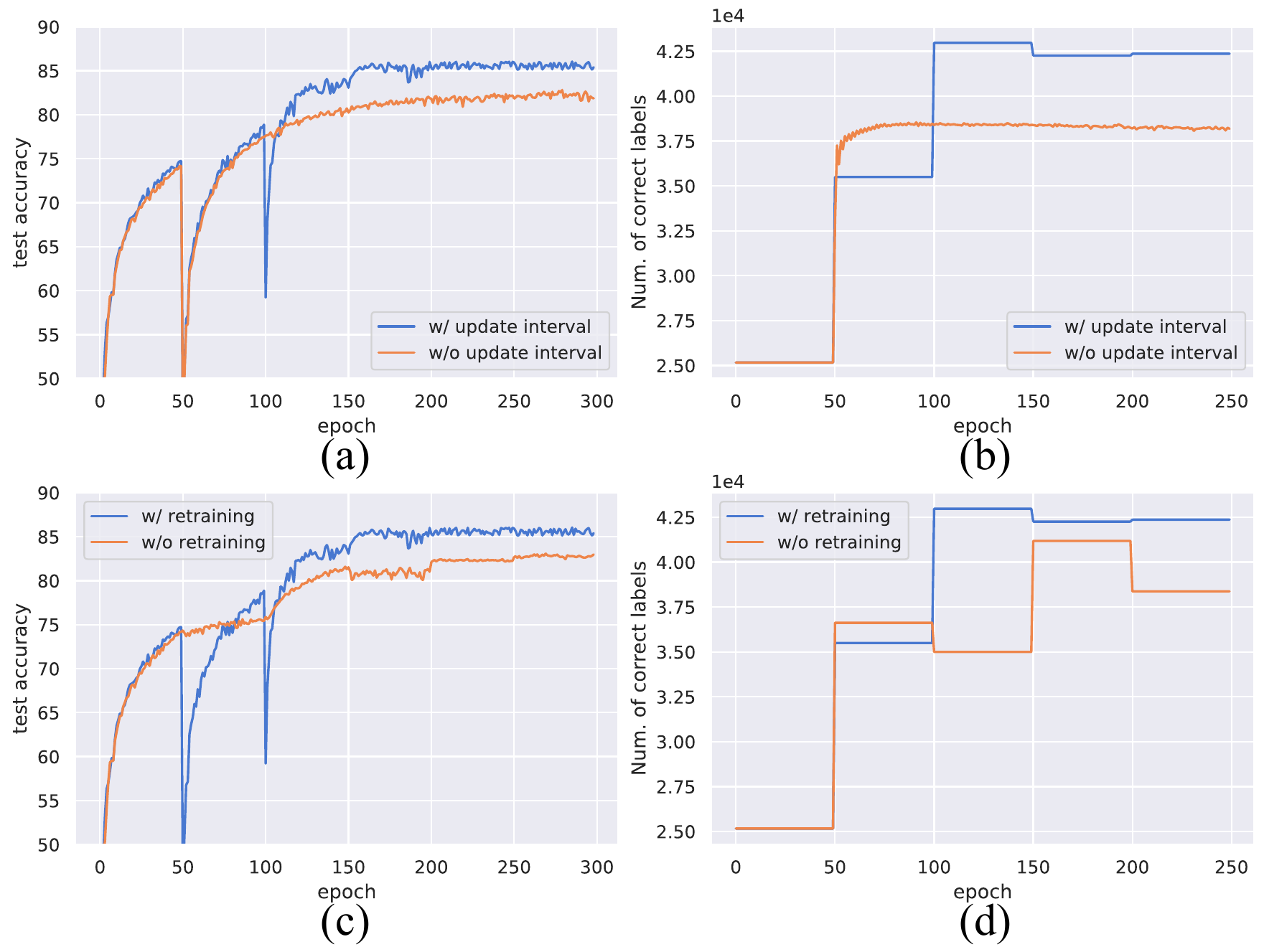}
\vspace{-0.4cm}
\caption{Comparison of ablation study over CIFAR-10. (a)(b): label update interval; (c)(d): retraining after correction.}
\label{fig:ab1}
\vspace{-0.6cm}
\end{figure}

Fig. \ref{fig:ab1} (a) and (b) show the test accuracy and number of correct labels versus epochs, respectively. In Fig. \ref{fig:ab1} (a), the large drop of the test accuracy indicates the trigger of retraining when the label correction rate is large. In continuous update, the label correction is smooth after the first correction. For fair comparison, we retrain networks with new labels after the first update.
We can see the test accuracy with update intervals is better than continuous update, and the number of correct labels is significantly higher than the latter. It proves the effectiveness of using relatively robust models to do the label correction.
Note that continuous correction has an average test accuracy of 82.1\%, which still maintains an advantage over the other methods in Table. \ref{tb:cifar10}.

{\bf Retraining After Correction}
Generally the network training with a fixed dataset is continuous for a large amount of epochs until it is stable. In Joint optimization and PENCIL \cite{tanaka2018joint,yi2019probabilistic} with label update, their labels and weights learning are performed in every iteration and the dataset will not be changed dramatically after each label correction. 
In terms of our method, the labels may change a lot in every label update, because the correction ability of networks after the label correction interval can be improved much. In our experiment, retraining the models with the updated dataset has a better performance than continuous training as shown in Fig. \ref{fig:ab1} (c)(d). 
In Fig. \ref{fig:ab1} (d), the number of correct labels without retraining is less than retraining, and it drops a lot in the last update. Correspondingly, the test accuracy of the former is lower than the latter. However, the mean accuracy of 82.8\% without retraining is still higher than that of other baselines.

\vspace{-0.6cm}
\section{Conclusion}
\vspace{-0.4cm}
We proposed a synergistic network learning and label correction methods to solve the noise label problem.
Our method is independent of the backbone network thus it is easy to apply. 
We demonstrated our method with synthetic label noise on CIFAR-10 and CIFAR-100 and real-world large scale dataset Clothing1M with label noise. Our method is an overall winner compared with the baselines.

\small
\bibliographystyle{IEEEbib}
\bibliography{strings,refs}

\end{document}